\documentclass[sigconf]{acmart}

\usepackage{balance}
\usepackage[T1]{fontenc}

\copyrightyear{2019}
\acmYear{2019}
\setcopyright{acmcopyright}
\acmConference[EMDL'19]{The 3rd International Workshop on Deep Learning for Mobile Systems and Applications}{June 21, 2019}{Seoul, Republic of Korea}
\acmPrice{15.00}
\acmDOI{10.1145/3325413.3329793}
\acmISBN{978-1-4503-6771-4/19/06}

\title{EmBench: Quantifying Performance Variations of\\ Deep Neural Networks across Modern Commodity Devices}

\author{
{Mario Almeida$^\dagger$*, Stefanos Laskaridis$^\dagger$*\\ Ilias Leontiadis$^\dagger$*, Stylianos I. Venieris$^\dagger$*, Nicholas D. Lane$^{\dagger,\ddagger}$}}
	
\affiliation{\institution{$^\dagger$Samsung AI Center, Cambridge\hspace{+0.75cm}$^\ddagger$University of Oxford}{\Small\textit{{* Indicates equal contribution.}}}}

\email{{mario.a, stefanos.l, i.leontiadis, s.venieris, nic.lane}@samsung.com}

\usepackage{natbib}
\usepackage{graphicx}
\usepackage{amsmath}
\usepackage{booktabs}
\usepackage{lipsum}
\usepackage{multirow}
\usepackage[flushleft]{threeparttable} 
\usepackage[colorinlistoftodos]{todonotes}
\usepackage{xspace}

\let\OldTexttrademark\texttrademark
\renewcommand{\texttrademark}{\OldTexttrademark\xspace}%

\usepackage{xcolor,soul}

\usepackage{background}
\backgroundsetup{angle=0,
    scale=1,
    color=black,
    firstpage=true,
    position=current page.north,
    hshift=0pt,
    vshift=-20pt,
    contents={\ifnum\value{page}=1 PREPRINT: Accepted at the 3rd International Workshop on Embedded and Mobile Deep Learning (EMDL), 2019 \else \fi}
}

\begin{abstract}
In recent years, advances in deep learning have resulted in unprecedented leaps in diverse tasks spanning from speech and object recognition to context awareness and health monitoring. As a result, an increasing number of AI-enabled applications are being developed targeting ubiquitous and mobile devices. While deep neural networks (DNNs) are getting bigger and more complex, they also impose a heavy computational and energy burden on the host devices, which has led to the integration of various specialized processors in commodity devices. Given the broad range of competing DNN architectures and the heterogeneity of the target hardware, there is an emerging need to understand the compatibility between DNN-platform pairs and the expected performance benefits on each platform. This work attempts to demystify this landscape by systematically evaluating a collection of state-of-the-art DNNs on a wide variety of commodity devices. In this respect, we identify potential bottlenecks in each architecture and provide important guidelines that can assist the community in the co-design of more efficient DNNs and accelerators.
\end{abstract}

\keywords{Deep neural networks; on-device inference; mobile devices}

\begin{document}

\maketitle

{\fontsize{8pt}{8pt} \selectfont
\textbf{ACM Reference Format:}\\
Mario Almeida, Stefanos Laskaridis, Ilias Leontiadis, Stylianos I. Venieris, Nicholas D. Lane. 2019. EmBench: Quantifying Performance Variations of Deep Neural Networks across Modern Commodity Devices. In \textit{ Proc. of The 3rd International Workshop on Deep Learning for Mobile Systems and
Applications, (EMDL'19), June 21, 2019, Seoul, Republic of Korea.} ACM, NY, NY, USA. 6 pages. DOI: https://doi.org/10.1145/3325413.3329793  }

\section{Introduction}
\label{sec:intro}

With a demonstrated state-of-the-art accuracy in a wide range of AI tasks, the popularity of deep neural networks (DNNs) is on the rise. Since 2012 and the introduction of \mbox{AlexNet \cite{krizhevsky2012imagenet}}, a myriad of models have been competing for improved predictive power \mbox{(Table \ref{tab:dnns})}. Nevertheless, accuracy gains have often been achieved at the expense of an increase in model complexity, inference time and resource requirements. With DNN models becoming ubiquitous across multiple scenarios and compute devices, from large-scale cloud \mbox{services \cite{microsoft2018}} to resource-constrained mobile systems \cite{Lane:2015:DLR:2699343.2699349}, predicting the processing performance of each DNN becomes a challenging task. Given the wide range of competing DNN architectures and the heterogeneity of the target hardware, there is an upcoming desire to gain insights on how and why different design decisions impact the accuracy and performance of these networks upon deployment.


\begin{table}[t]
    \centering
    \resizebox{\linewidth}{!}{
        \begin{tabular}{lllllll}
        \toprule
        Model & Year & FLOPs (M) & Params (M) & Accuracy  & Accuracy \\ 
              &      &           &            & (top-1)   & (top-5) \\
        \midrule
        AlexNet \cite{krizhevsky2012imagenet}                              & 2012 & 955.21   & 61.10  & 56.52   & 79.07  \\
        VGG 11 \cite{simonyan2014very}                                     & 2014 & 8171.57  & 132.86 & 69.02   & 88.63  \\
        VGG 13 \cite{simonyan2014very}                                     & 2014 & 11895.04 & 133.05 & 69.93   & 89.25  \\
        VGG 16 \cite{simonyan2014very}                                     & 2014 & 16063.36 & 138.36 & 71.59   & 90.38  \\
        VGG 19 \cite{simonyan2014very}                                     & 2014 & 20231.68 & 143.67 & 72.38   & 90.88  \\
        ResNet 18 \cite{He2015}                                            & 2015 & 1836.82  & 11.69  & 69.76   & 89.08  \\
        ResNet 34 \cite{He2015}                                            & 2015 & 3692.78  & 21.80  & 73.31   & 91.42  \\
        ResNet 50 \cite{He2015}                                            & 2015 & 4154.96  & 25.56  & 76.13   & 92.86  \\
        ResNet 101 \cite{He2015}                                           & 2015 & 7892.77  & 44.55  & 77.37   & 93.55  \\
        ResNet 152 \cite{He2015}                                           & 2015 & 11636.60 & 60.19  & 78.31   & 94.05  \\
        Inception-v3 \cite{szegedy2017inception}                           & 2015 & 5730.17  & 27.16  & 75.64   & 92.59  \\
        Inception-v4 \cite{szegedy2017inception}                           & 2015 & 12561.10 & 42.68  & 80.08   & 94.89  \\
        SqueezeNet-v1 \cite{iandola2016squeezenet}                         & 2016 & 865.78   & 1.25   & 58.09   & 80.42  \\
        SqueezeNet-v1.1 \cite{iandola2016squeezenet}                       & 2016 & 377.80   & 1.24   & 58.18   & 80.62  \\
        DenseNet 121 \cite{huang2017densely}                               & 2017 & 2928.89  & 7.98   & 74.43   & 91.97  \\
        DenseNet 169 \cite{huang2017densely}                               & 2017 & 3473.88  & 14.15  & 75.60   & 92.81  \\
        DenseNet 201 \cite{huang2017densely}                               & 2017 & 4435.03  & 20.01  & 76.87   & 93.37  \\
        DenseNet 161 \cite{huang2017densely}                               & 2017 & 7902.37  & 28.68  & 77.14   & 93.56  \\
        Xception \cite{chollet2017xception}                                & 2017 & 8494.59  & 22.86  & 78.89   & 94.29  \\
        MobileNetV2 \cite{mobilenetv2_2018}                                & 2017 & 336.43   & 3.50   & 71.81   & 90.42  \\
        ShuffleNet-v2.05 \cite{ma2018shufflenet}      & 2018 & 52.32    & 1.37   & 60.55   & 81.75  \\
        ShuffleNet-v2.1 \cite{ma2018shufflenet}       & 2018 & 160.09   & 2.28   & 69.36   & 88.32  \\
        MnasNet \cite{tan2018mnasnet}                                      & 2018 & 649.51   & 4.38   & 61.95   & 84.73  \\
        PNASNet \cite{liu2018progressive}                                  & 2018 & 25945.87 & 86.06  & 82.74   & 95.99  \\
        NasNet \cite{zoph2018learning}                                     & 2018 & 24882.21 & 88.75  & 82.51   & 96.02  \\
        NasNet mobile \cite{zoph2018learning}                              & 2018 & 667.75   & 5.29   & 74.08   & 91.74  \\
        \bottomrule
        
        \end{tabular}
    }
    \vspace{0.1cm}
    \caption{DNN benchmarks.}
    \label{tab:dnns}
    \vspace{-0.75cm}
\end{table}

EmBench aims to provide a comprehensive analysis of widely used deep neural networks, with a focus on evaluating which models thrive under which target platforms, while identifying their bottlenecks and sources of inefficiency. 
To this end, we analyze a set of popular DNNs (Table \ref{tab:dnns}) targeting various compute platforms (Table \ref{tab:platforms}).
First, we perform a macro analysis of these networks in terms of their complexity, inference latency and throughput for multiple batch sizes.
Then, we perform a deeper analysis of the most prominent network operations, with a focus on detecting non-trivial differences between execution across the target platforms. 

More specifically, we provide the following contributions:
\begin{itemize}
    \item We demonstrate that different networks are handled quite differently by each target platform, making it challenging to design an efficient model in a hardware-agnostic manner.
    \item We provide insights about how different batch sizes can affect performance on five different hardware architectures.
    \item We analyze the inference latency of all networks and show that the trade-off between actual processing speed and accuracy depends on the underlying hardware and its \mbox{optimizations}.
    \item We break down the overall DNN workload into individual operations and unveil any opportunities for further improvements on each platform. 
\end{itemize}

\section{Related Work}
\label{sec:related_work}

So far, a few studies have focused on analyzing the system-level properties of DNNs on deployment platforms. Canziani \textit{et al.} \cite{canziani2016analysis} presented a system-level analysis of 14 convolutional neural networks (CNNs) on the NVIDIA Jetson TX1 platform. 
Despite the fact that the analysis spanned across multiple metrics, the study was conducted over a limited number of networks and targeted solely a single platform. 
Bianco \textit{et al.} \cite{bianco2018benchmark} extended the covered space by evaluating a wider range of networks and targeting one embedded (Jetson TX1) and one high-end compute platform (NVIDIA Titan X GPU). 
Both studies conducted an analysis of the selected networks across multiple dimensions, including accuracy, compute speed, memory footprint and power consumption. 
Nevertheless, by including a total of two platforms --and given the heterogeneity of currently available devices-- the presented insights are not directly transferable to platforms with different characteristics.

In this work, we expand to a broad range of both high- and low-end devices, spanning from the latest server-grade RTX 2080 Ti GPU and the embedded Nvidia Jetson AGX down to the mobile-ready Qualcomm mobile Kryo 385 CPU and the low-power Intel Neural Compute Stick 2.
Furthermore, we present a microscopic view of how well different layer types are mapped to each hardware architecture, aiming to provide insights for the hardware-aware design of novel DNNs.

On a slightly different setting, Huang \textit{et al.} \cite{huang2017speed} concentrated on the task of object detection and evaluated a wide set of CNN-based object detectors in terms of processing performance and detection accuracy. 
With a focus on the mobile space, Ignatov \textit{et al.} \cite{Ignatov_2018_ECCV_Workshops} assembled a benchmark suite of representative AI tasks to assess the processing capabilities of currently available smartphones. 
In this paper, we adopt a wider scope than \cite{huang2017speed} by treating network architectures in a task-agnostic manner and target more diverse families of devices compared to \cite{Ignatov_2018_ECCV_Workshops}. 

Last, Zhang \textit{et al.} \cite{Zhang2018} study the key performance differences among different machine learning frameworks across different edge platforms. We treat the framework as an invariant and focus our endeavours on the inference behaviour of the devices at hand for a significantly greater variety of models.

\section{Hardware Platforms}
\label{sec:hw_accel}

\begin{table}[t]
\centering
\resizebox{\linewidth}{!}{
    \begin{threeparttable}
        \begin{tabular}{l c c c c l l}
        \toprule
        \multicolumn{1}{l}{\textbf{Platform}} & \textbf{Cores} & \begin{tabular}[l]{@{}c@{}} \textbf{Clock Freq.} \\ \textbf{(GHz)} \end{tabular} & \begin{tabular}[l]{@{}c@{}} \textbf{Memory} \\ \textbf{(GB)} \end{tabular} & \begin{tabular}[l]{@{}c@{}} \textbf{Technology} \\ \textbf{(nm)} \end{tabular} & \begin{tabular}[l]{@{}c@{}} \textbf{TDP} \\ \textbf{(W)} \end{tabular} \\ 
        
        \midrule
        Intel Xeon 4116* & \phantom{0}12 & 2.1 & 256 & 14 & \phantom{0}85 \\
        NVIDIA RTX 2080 Ti & 512$^\dagger$ & 1.5 & \phantom{0}11 & 12 & 250 \\
        Qualcomm Kryo 385 CPU & 4+4** & 2.8 + 1.8 & \phantom{00}6 & 10 & \phantom{00}5 \\
        Intel NCS 2 & 16$^\ddagger$ & 0.7 & 0.5 & 16 & \phantom{00}1 \\ 
        NVIDIA Tegra Xavier GPU & 512$^\dagger$ & 1.3 & \phantom{0}16 & 12 & \phantom{0}30 \\
        \bottomrule
        
        \end{tabular}
        \begin{tablenotes}
            \footnotesize
            \item * HyperThreading enabled
            \item ** 4 high-performance ARM A75 + 4 high-efficiency ARM A55. 
            \item $\ddagger$ Movidius SHAVE cores. $\dagger$ 512 CUDA cores w/ 64 Tensor Cores.
        \end{tablenotes}
    \end{threeparttable}
}
\vspace{0.05cm}
\caption{Evaluated platforms.}
\label{tab:platforms}
\vspace{-1cm}
\end{table}

The large compute and memory demands of modern DNN workloads have led to an emergence of specialized processors with the goal to facilitate their high-performance deployment. Depending on the target scenario, each platform has employed different hardware optimizations to satisfy system-level constraints, including latency, throughput, temperature, power dissipation and form factor. 

In desktop and server environments, high-end devices are typically employed in order to maximize throughput at the penalty of substantial power consumption. In this context, powerful --and massively parallel-- but power-costly platforms have been designed. A representative example is the latest NVIDIA GeForce RTX 2080 Ti GPU which is based on the NVIDIA Turing architecture. By introducing the 2nd generation of Tensor Cores --a set of specialized hardware units tailored for DNN processing-- this particular GPU provides hardware support for 16-bit floating-point as well as 8- and 4-bit fixed-point precision and enables the highly optimized execution of matrix operations with mixed-precision arithmetic. Having a large bandwidth of 616 GB/s to a 11 GB off-chip RAM, the RTX 2080 Ti GPU has been optimized for high throughput, reaching its peak performance when processing inputs in batches, yielding up to 13.4 (FP32) TFLOP/s at the cost of a 250-watt peak power.

On the other end of the spectrum, severely power-constrained systems such as IoT and mobile devices are equipped with processing units that comply with the respective thermal and form-factor constraints. 
In this space, low-power miniaturized SoCs, such as 
the Qualcomm Snapdragon\texttrademark 845 (SDM845), have been explicitly designed with the objective to provide the processing support for on-device DNNs while respecting the typical <10 watts power budget of modern consumer and robotic systems. 
Further towards customization, neural accelerators offer sub-watt solutions for rapid DNN inference in severely constrained embedded systems. 
To this end, full-custom chips have been designed with the goal to extract the highest possible performance at a minimal power dissipation. 
A representative and widely available instance is the Intel Neural Compute Stick (NCS) 2 mounting the Movidius Myriad X accelerator, delivering up to 1 TOP/s under 1 watt.

Targeting a mid-range power envelope of a few 10s of watts which is typical in complex embedded systems such as driverless cars and robots, several devices have been designed to reach a configurable trade-off between power consumption and processing speed. A representative example is the latest NVIDIA Jetson Xavier SoC. 
Jetson Xavier hosts a 512-core NVIDIA Volta GPU with 64 1st generation Tensor Cores. To support different power budgets while sustaining a high performance, the platform can be configured with a range of frequencies up to 1.3 GHz at a peak TDP of 30 watts.

Given the diversity of DNN-enabled applications, the variability across deployment scenarios, network architectures and hardware platforms leads to an emerging need for evaluating the compatibility between different network-platform pairs. 
In this respect, Section \ref{sec:evaluation} examines the efficiency of mapping a substantially comprehensive range of networks (Table \ref{tab:dnns}) on each platform of Table \ref{tab:platforms} with the potential to guide both novel network and hardware designs.


        
        

\begin{figure}[t]
\centering
\includegraphics[width=0.475\textwidth]{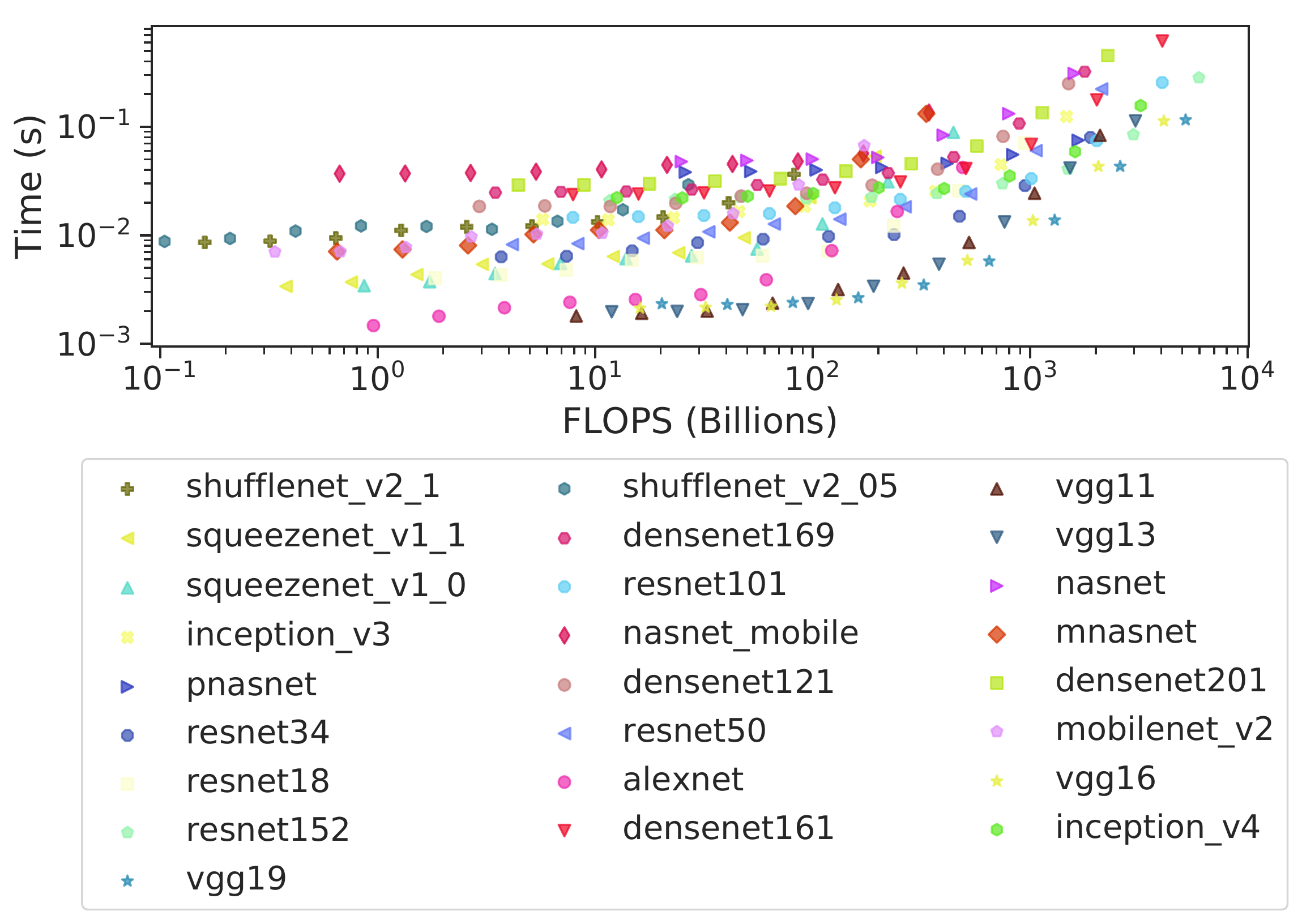}
\vspace{-0.2cm}
\caption{Runtime vs. number of FLOPs on a single NVIDIA RTX 2080 Ti GPU. Different markers represent different networks. Each network has an increasing amount of FLOPs based on the batch size (\textit{i.e.}, FLOPs/inference $\times$ batch size).}
\label{fig:flops}
\end{figure}

\begin{figure}[t]
    \centering
    \includegraphics[width=0.475\textwidth]{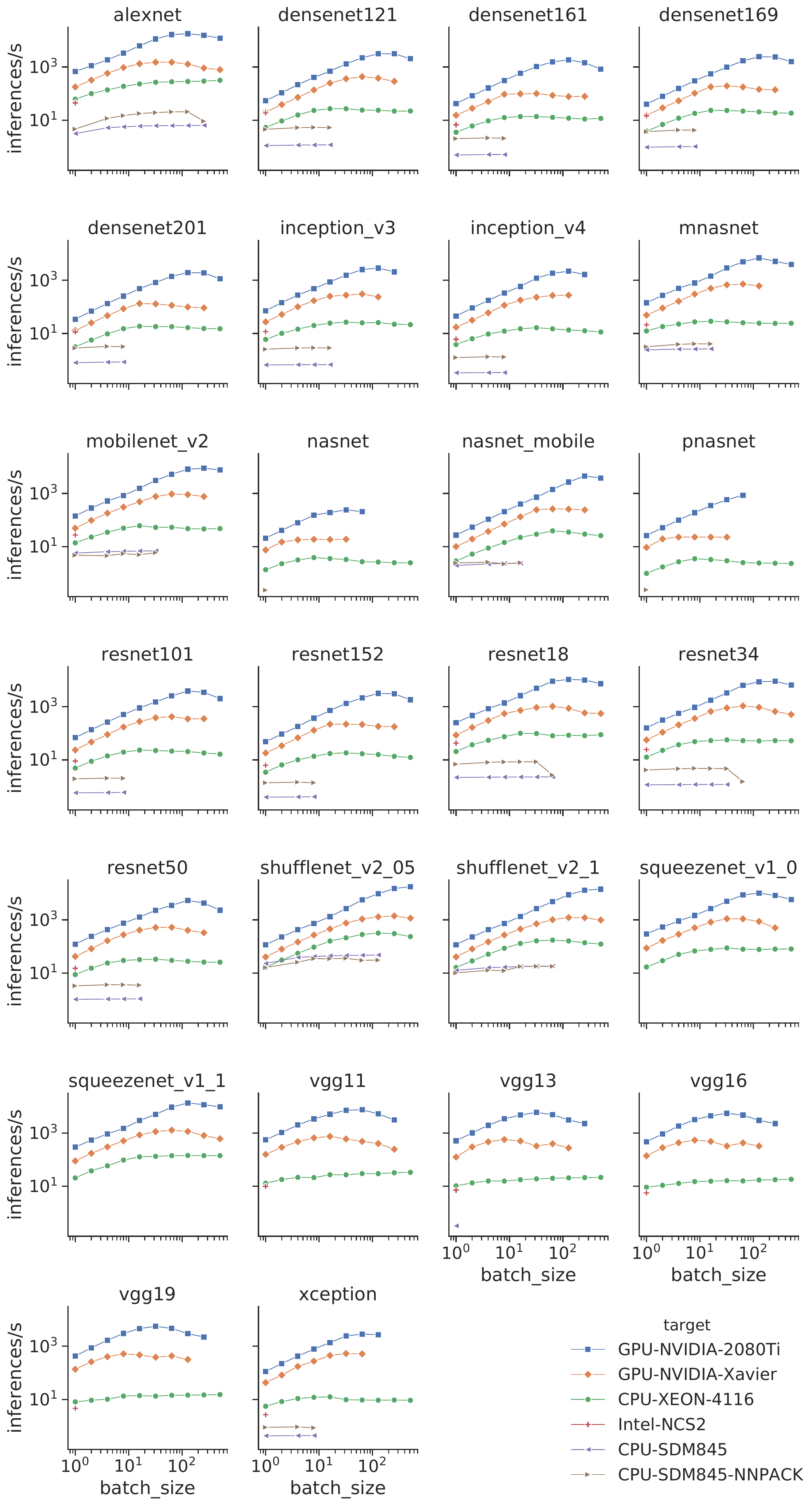}
    \vspace{-0.2cm}
    \caption{Achieved throughput of evaluated DNNs on different platforms with increasing batch size.}
    \label{fig:inferences}

\end{figure}

\section{Evaluation}
\label{sec:evaluation}

\begin{figure*}[ht]
\centering
    \centering
        \includegraphics[width=0.33\textwidth]{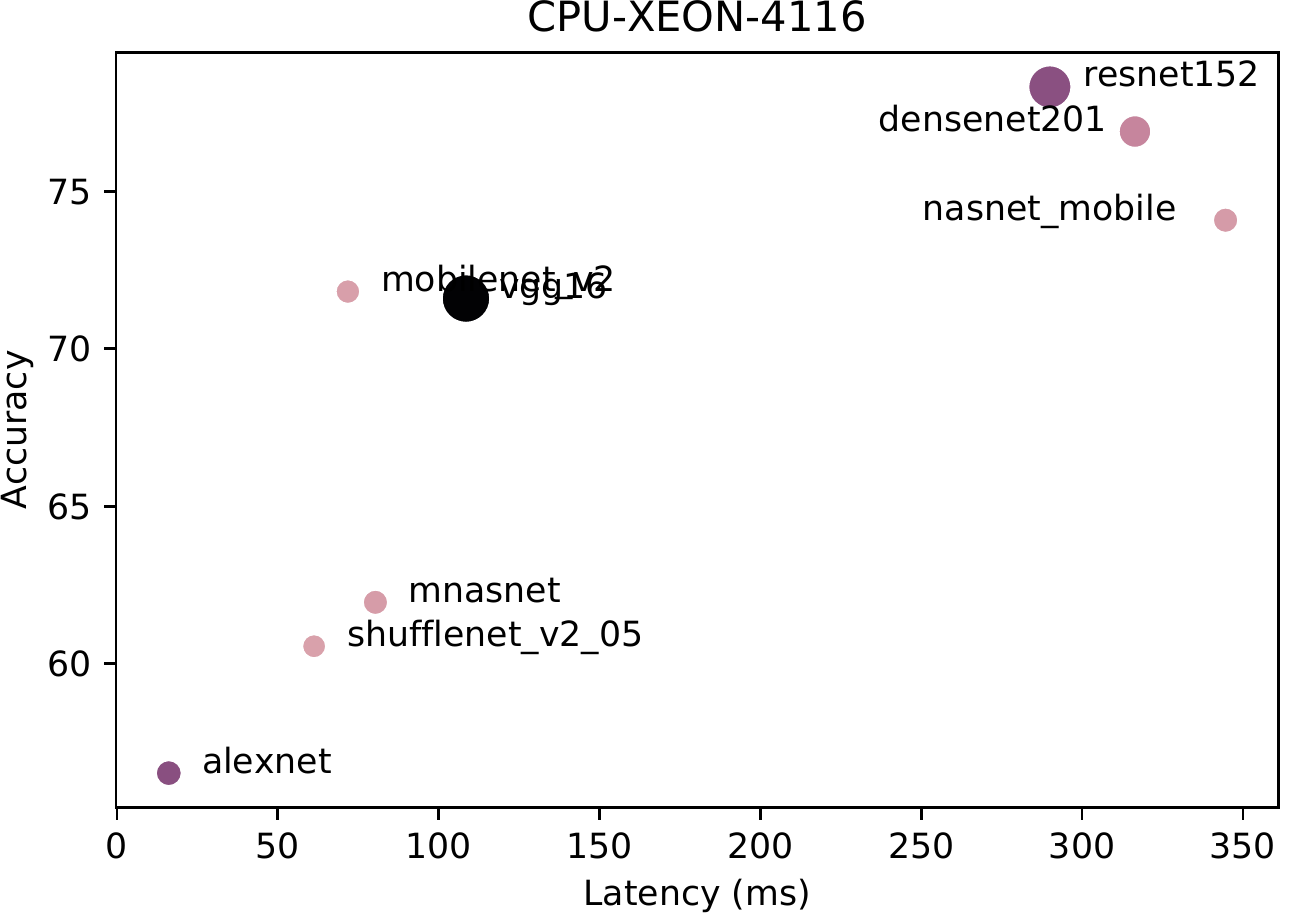}
        \hfill
        \includegraphics[width=0.33\textwidth]{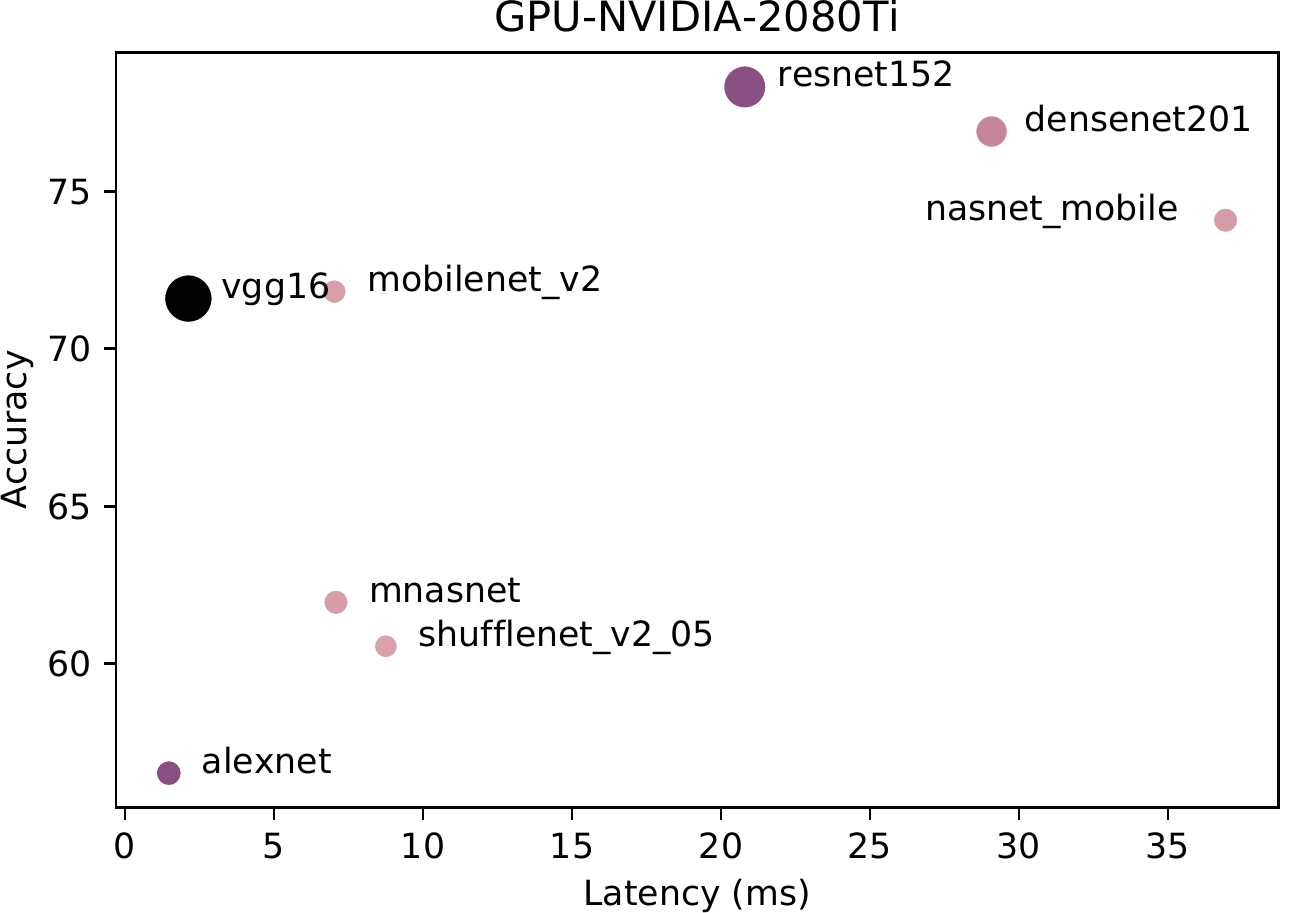}
        \hfill
        \includegraphics[width=0.33\textwidth]{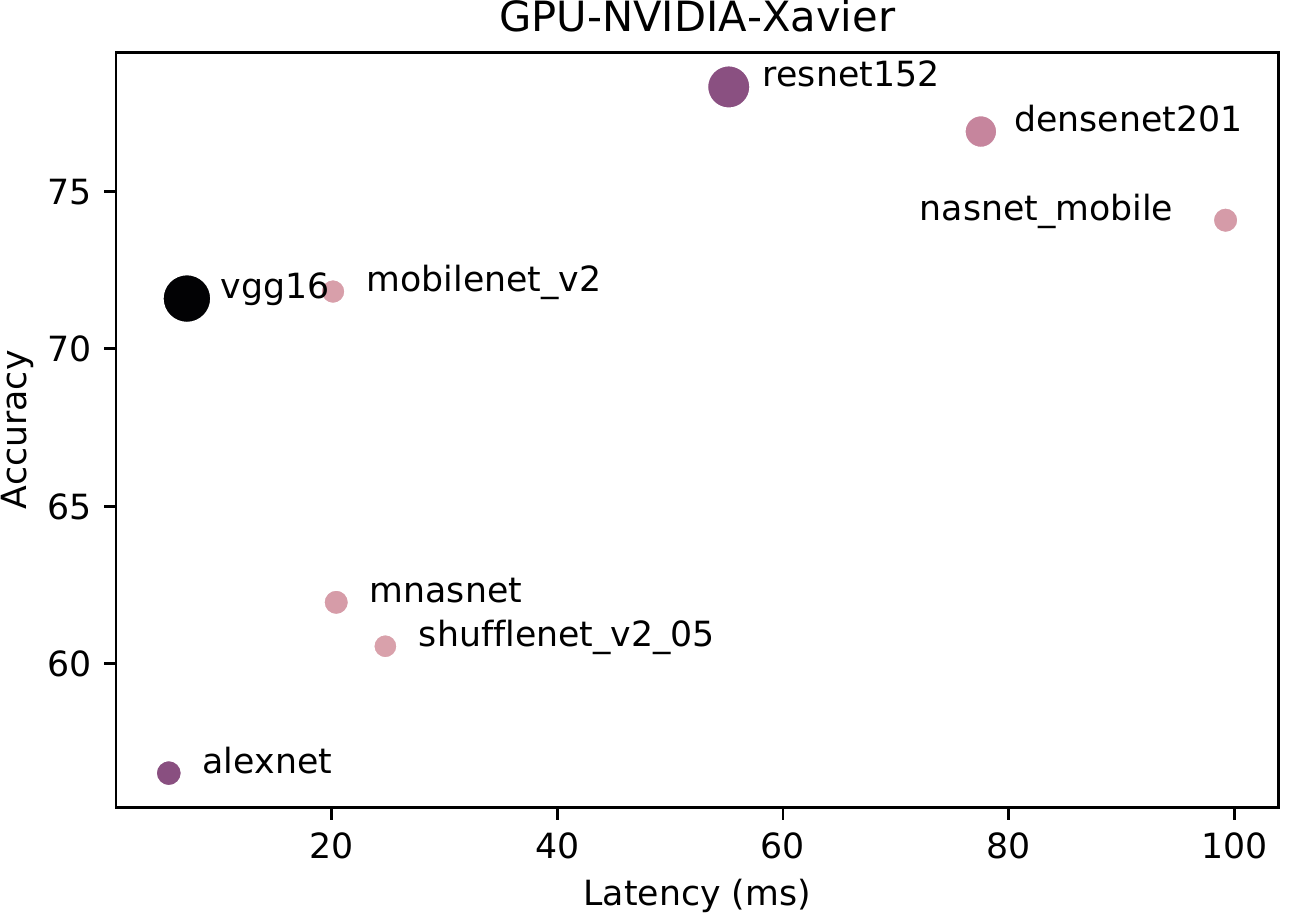}\\[3ex]
        \includegraphics[width=0.33\textwidth]{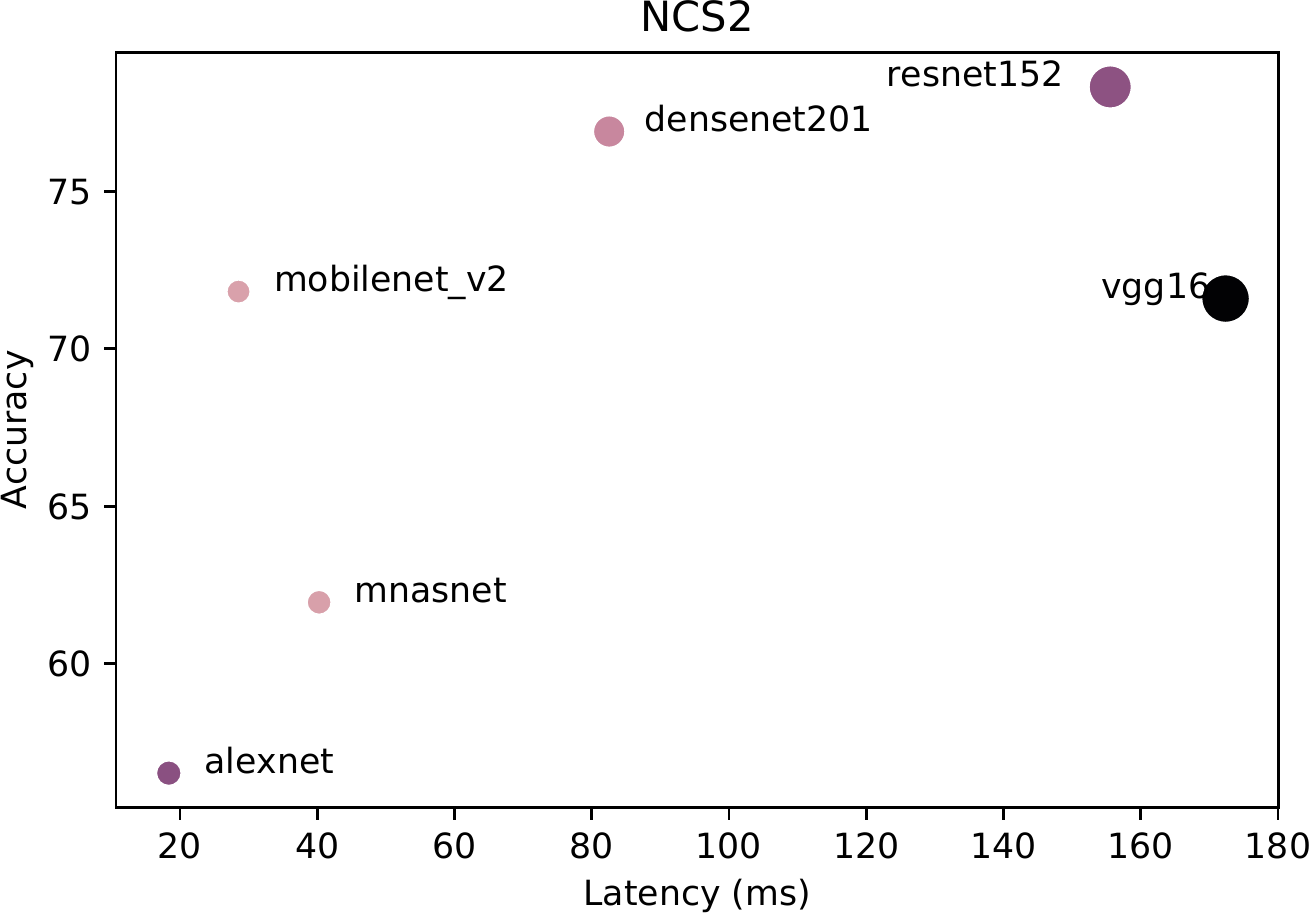}
        \hfill
        \includegraphics[width=0.33\textwidth]{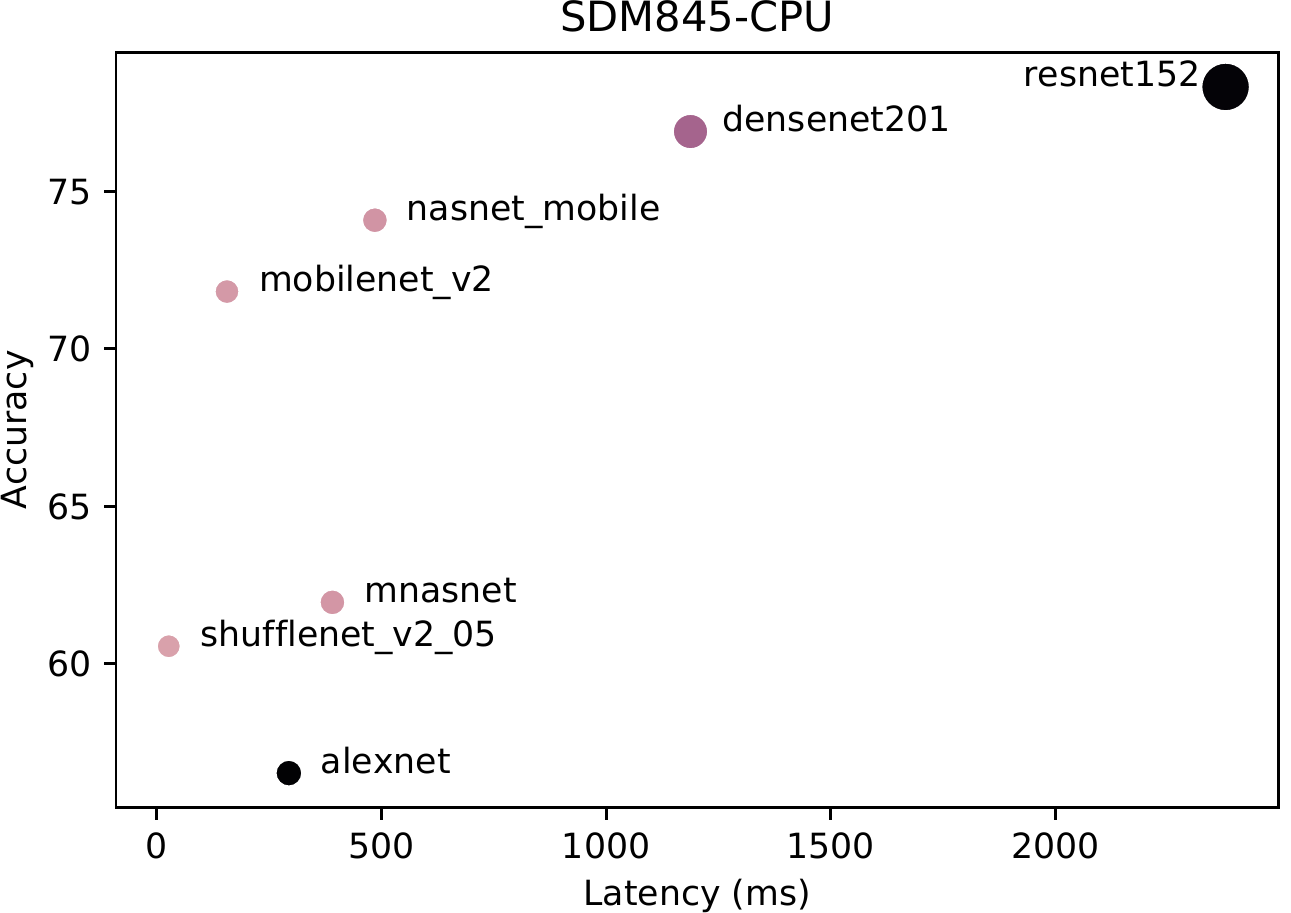}
        \hfill
        \includegraphics[width=0.33\textwidth]{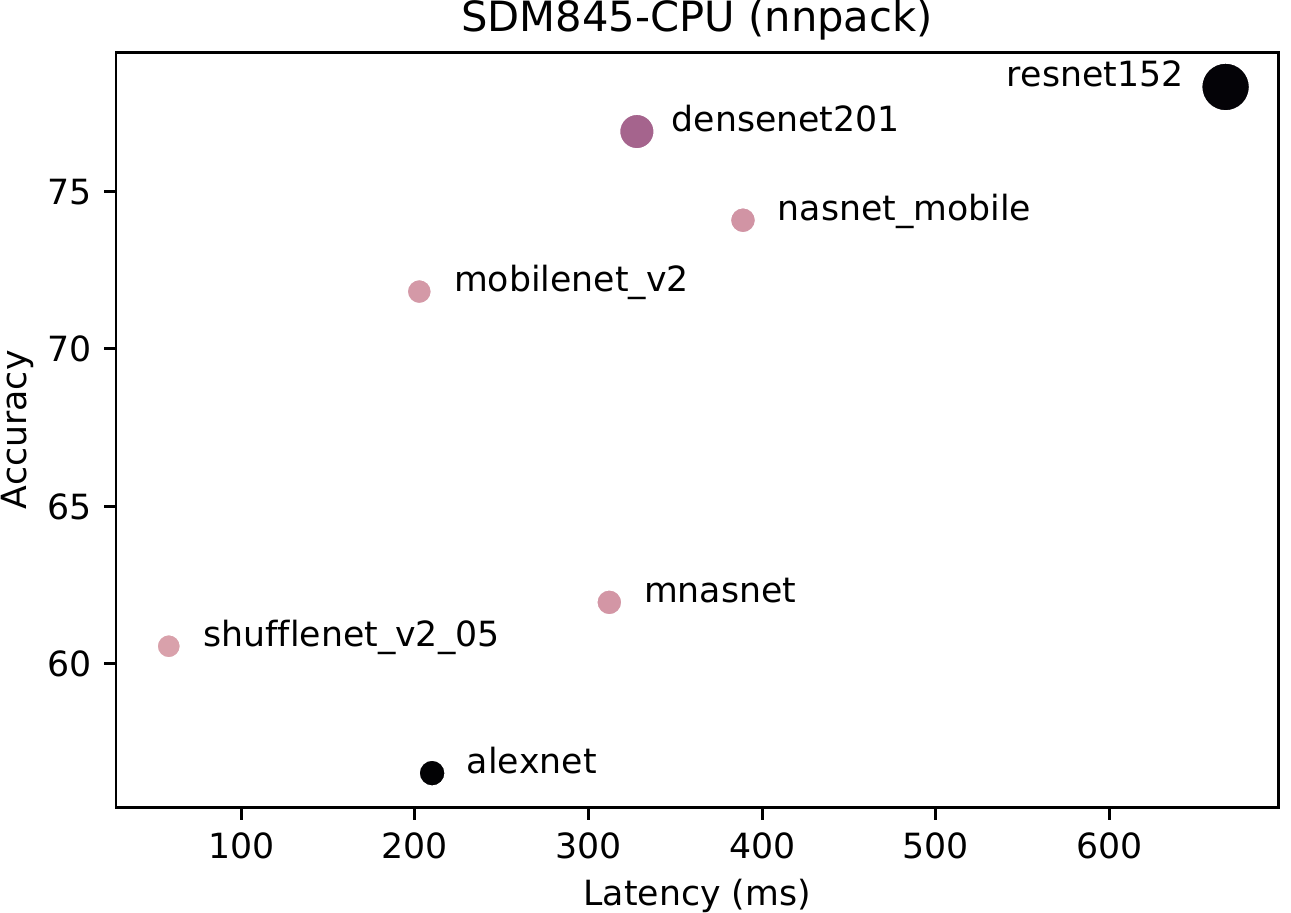}
\caption{Inferences-per-second vs. accuracy for various target platforms (batch size = 1). The marker size and color represent the number of FLOPs and parameters, respectively, in each of the evaluated networks.}
\label{fig:inf_per_acc}

\end{figure*}

In this section, we analyze the data that we have collected by benchmarking the set of networks from Table \ref{tab:dnns} targeting various commodity compute platforms (Table \ref{tab:platforms}). Each configuration comprises a choice of i) DNN, ii) batch size and iii) target platform. For each configuration, we initially perform a macroscopic analysis with respect to complexity, measured in number of floating-point operations, and achieved latency on the selected device. We further analyze the latency of each operation type within the network to identify processing bottlenecks and compare the computational efficiency across the target devices.


\textbf{Benchmarking procedure.}
Following the effort of the Open Neural Exchange Format (ONNX) to provide a uniform interface among deep-learning frameworks, we adopted Facebook's machine-learning toolchain (PyTorch, Caffe2, FAI-PEP) for the majority of our experiments due to its imperative interface and support for ONNX. 
First, the pretrained versions of all DNNs were collected in a PyTorch format.
Specifically, for the experiments on a workstation hosting the Xeon 4116 CPU and RTX 2080 Ti GPU, and on the NVIDIA Jetson Xavier SoC, PyTorch\footnote{https://pytorch.org/} v1.0.0 with CUDA 10 and cuDNN 7.5 are used directly. 
The two platforms run GNU/Linux Ubuntu 18.04 LTS, compiled for x86-64 and 64-bit ARM, respectively.
For the Xavier SoC, we set the GPU frequency to its maximum setting to obtain the peak performance.
On the mobile side, the evaluated networks were converted to Caffe2 to run on SDM845's Kryo CPU, running Android 9 (Pie). 
The Caffe2 backend for mobile platforms was configured to employ the highly optimized implementations of the NNPACK\footnote{https://github.com/Maratyszcza/NNPACK} package.
To systematically measure on-device performance, we employ Facebook's AI Performance Evaluation Platform\footnote{https://github.com/facebook/FAI-PEP} (FAI-PEP).
Finally, when targeting Intel NCS 2, the networks were converted to ONNX and subsequently to the Intel Movidius graph file format through the Intel OpenVINO\texttrademark toolkit. 
To time the execution of DNNs on NCS 2, we use the program counters of the Myriad X chip.

Across all platforms, each experiment is run 50 times -- with ten warm-up runs for uniform initial cache state and a cool-off period of 2 seconds between runs to avoid frequency scaling due to heat -- and the minimum latency achieved is reported.


\textbf{FLOPs analysis.} Popular DNNs tend to vary quite a lot in terms of their complexity.
As seen on Table \ref{tab:dnns}, some networks require up to 3 orders of magnitude more floating-point operations (FLOPs) to perform image recognition over the same dataset (\textit{e.g.}, \mbox{ImageNet \cite{Fei-Fei2010}}), often with little to no improvement in accuracy, such as in the case of MobileNetV2 vs. VGG16.

We first investigate if the FLOPs are a good indicator of inference time. 
Figure \ref{fig:flops} shows how the actual inference time on the 2080 Ti GPU varies with respect to FLOPs.
The amount of FLOPs on the x-axis is the network's FLOPs times the batch size.
We used powers of two for the batch sizes up to 512.
We can see that for networks with similar FLOPs, the actual GPU time can vary by at least one order of magnitude. Nonetheless, for the same network, the GPU time grows slowly with the batch size up to a point where time growth becomes almost exponential. At this point, the GPU resources are fully utilized and the cost of further batching is no longer amortized.

\textbf{Impact of batch size.}
Figure \ref{fig:inferences} depicts the effect of batch size on the achieved throughput across models and platforms. On the one side of the spectrum, given the memory-abundant workstation setup, the highest performing batch sizes for the high-end 2080 Ti GPU typically lie between 128 and 256. At this point, the GPU reaches the peak utilization of its resources and, consequently, after that the overhead of further batching is no longer amortized. In this case, the GPU is able to exploit the intra-batch parallelism of large batches and boost the achieved throughput, while not being limited by the available RAM. On the other hand, although similarly to the high-end GPU there are no significant memory constraints, the Xeon 4116 CPU rarely improves its throughput after a batch size of 32, where typically its resources --which are less than the resource-rich GPU's-- are already fully utilized. 

On the other side of the spectrum, large batches tend to exhaust the memory-limited devices. 
In this respect, the mobile Kryo 385 CPU (CPU-SDM845 in Figure \ref{fig:inferences}) 
shows modest throughput improvement for increasing batch size due to both the reduced memory bandwidth and amount of compute resources compared to its high-end counterparts, and its maximum batch size is typically up to 32 due to the small off-chip memory capacity.
Note that using NNPACK's CPU optimized convolution layers (CPU-SDM845-NNPACK in Figure \ref{fig:inferences}) improves the number of inferences per second by $2.97\times$ on average and up to $4.53\times$ for densenet121.

Moreover, NCS 2 outperforms its mobile counterpart across the evaluated networks, despite the fact that it was designed to operate with a batch size of 1. 
Finally, the embedded Tegra Xavier GPU demonstrates a similar pattern to RTX 2080 Ti by being able to exploit batch processing to improve its throughput, although its median optimal batch size is around 64.
The highest throughput is observed at the point where its resources are fully utilized, after which further batching degrades the achieved throughput due to excessive uncompensated overhead.


\begin{figure*}[ht]
\centering
    \centering
        \includegraphics[width=0.31\textwidth]{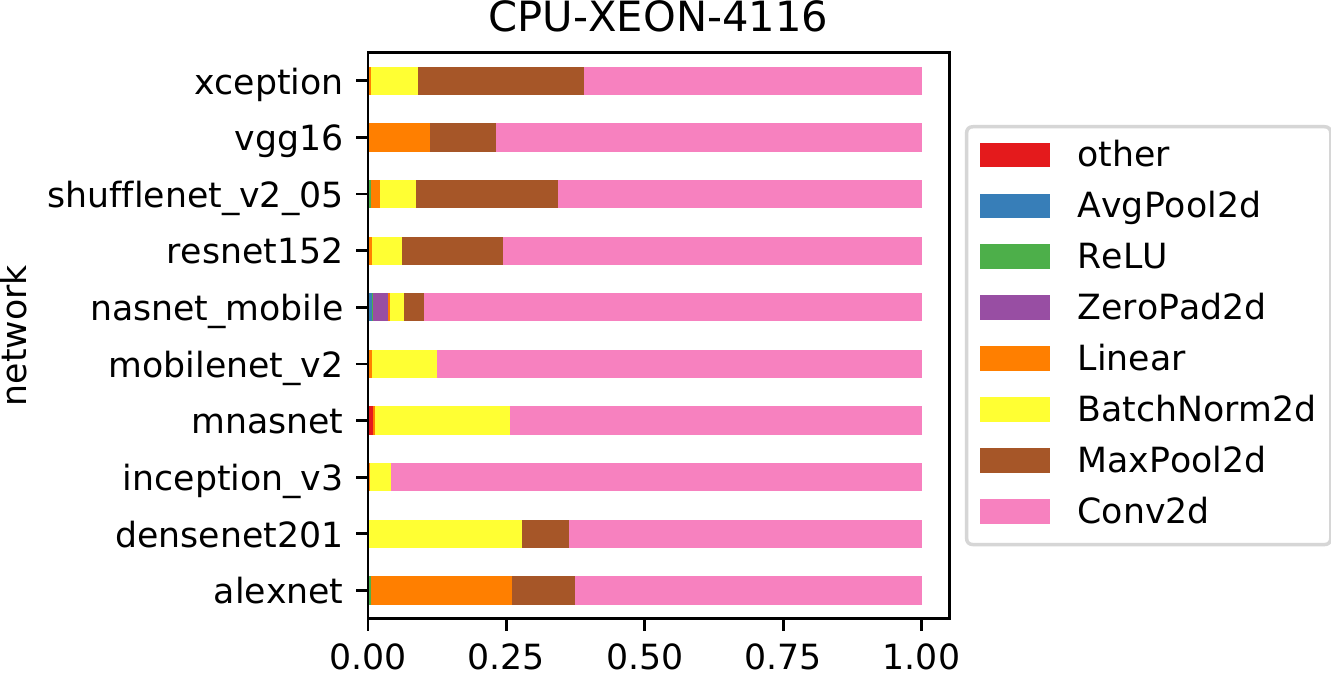}
        \hfill
        \includegraphics[width=0.33\textwidth]{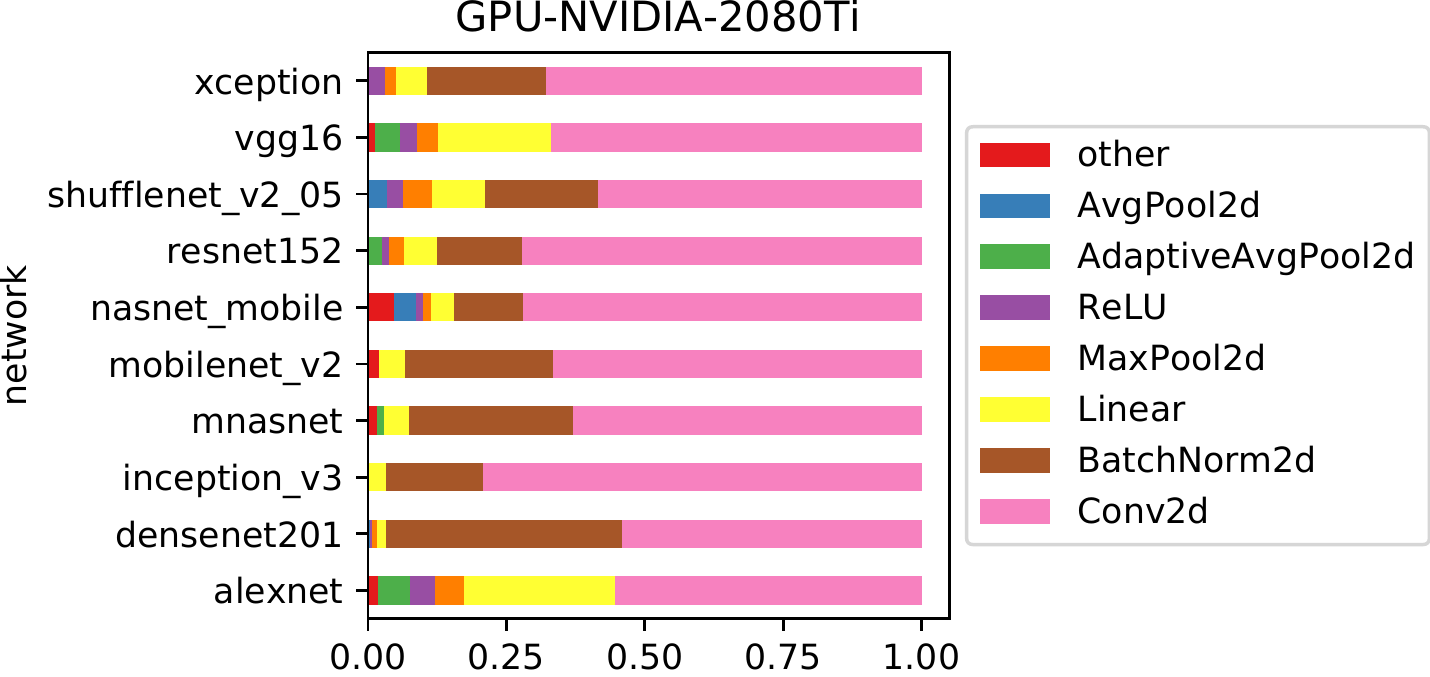}
        \hfill
        \includegraphics[width=0.31\textwidth]{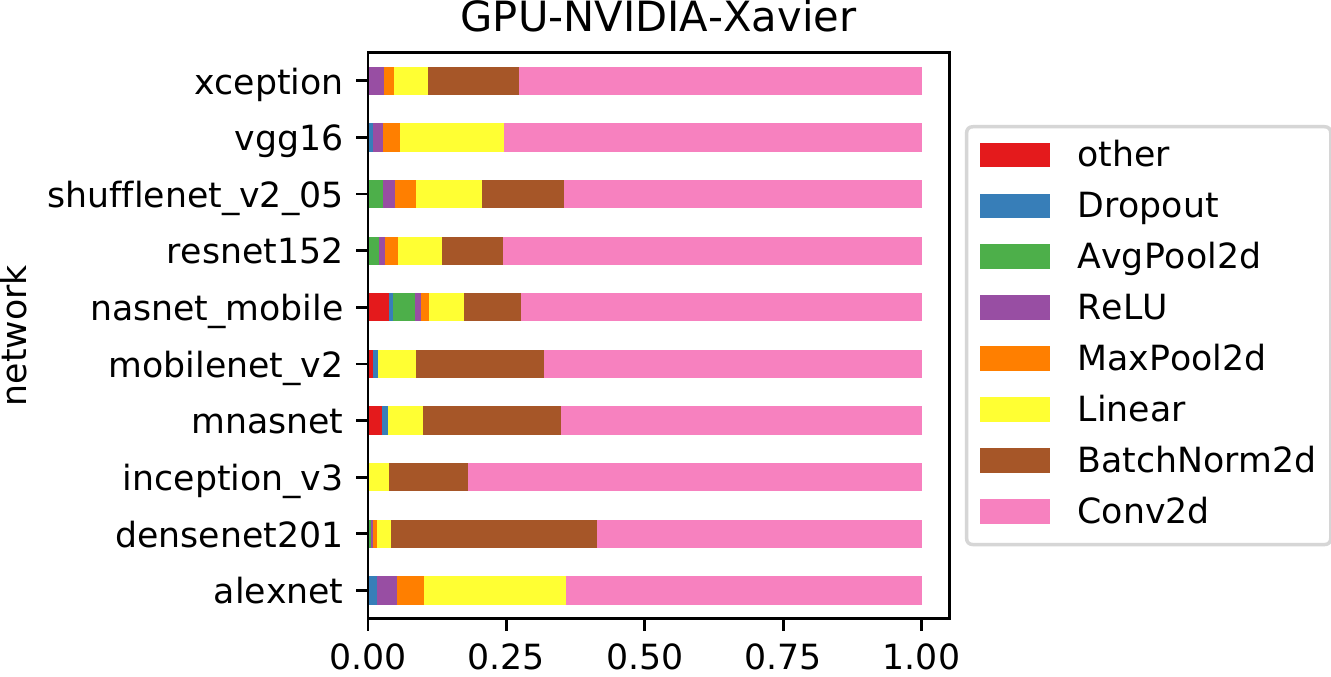}\\[2ex]
        \includegraphics[width=0.31\textwidth]{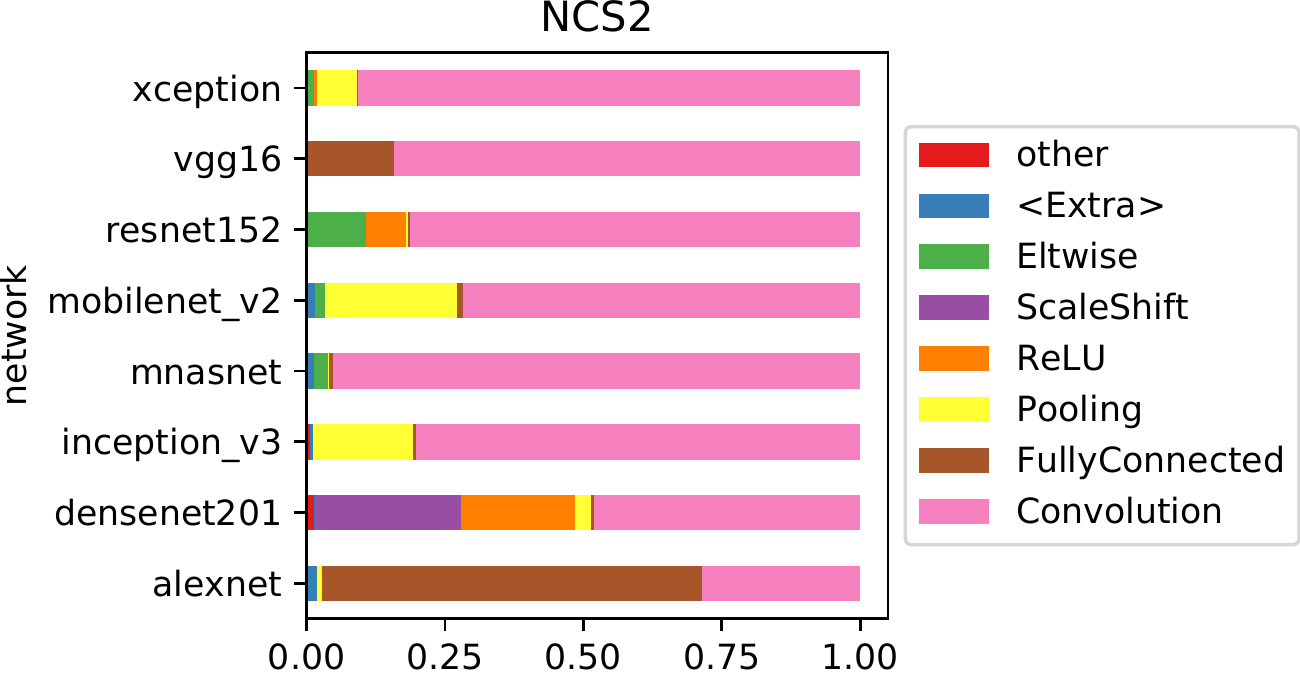}
        \hfill
        \includegraphics[width=0.31\textwidth]{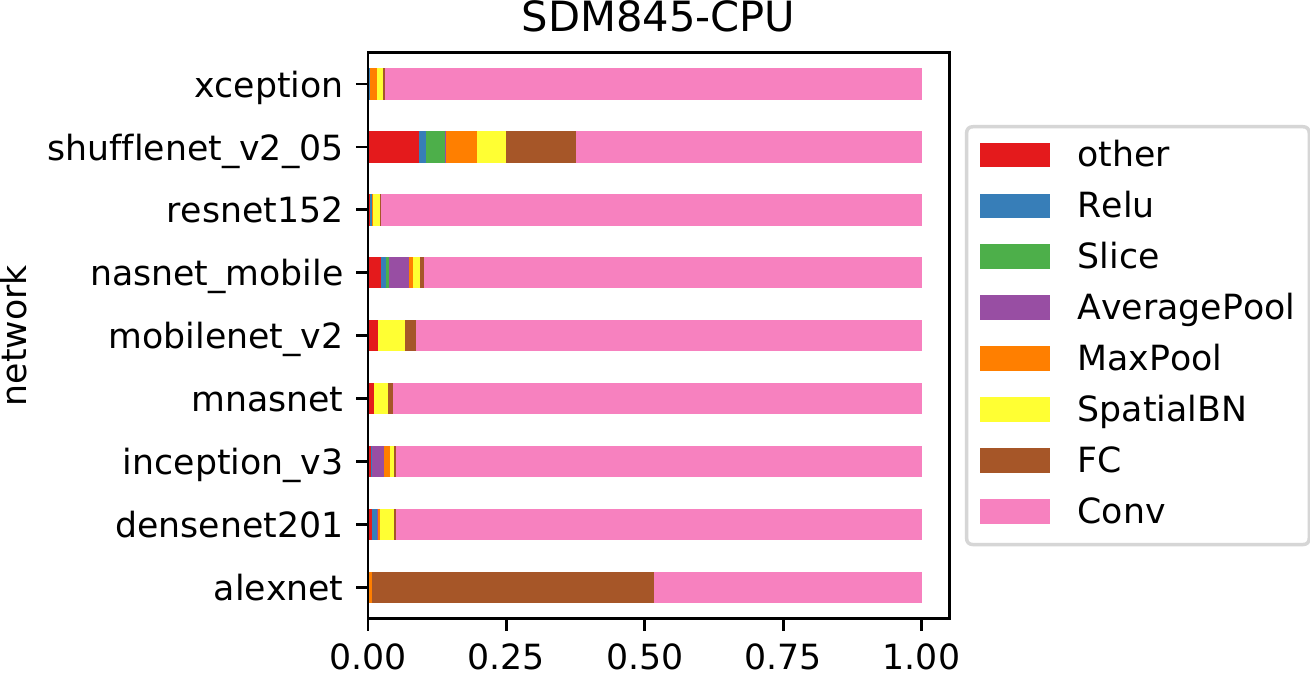}
        \hfill
        \includegraphics[width=0.31\textwidth]{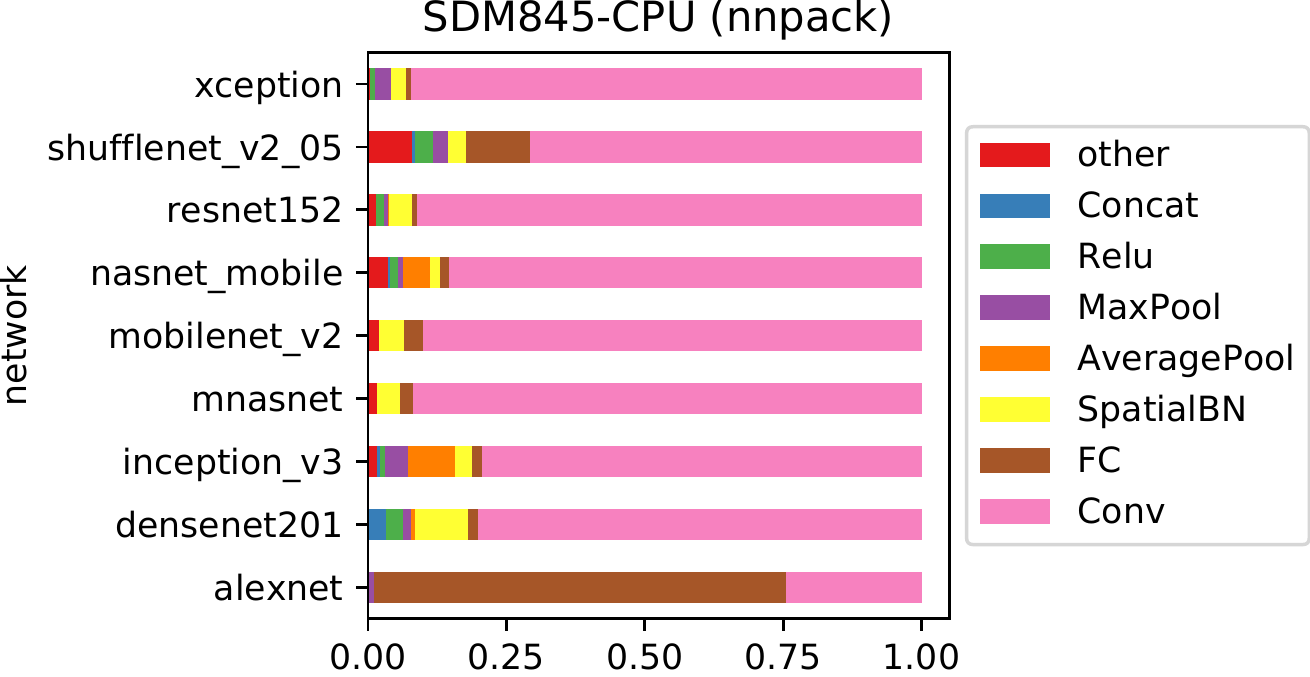}

\caption{Percentage of time spent per layer for various target platforms (batch size = 1). Only a sample of networks is shown.}
\label{fig:per_layer}
\end{figure*}  
    



\textbf{Accuracy vs. time.}
Figure \ref{fig:inf_per_acc} shows the accuracy and achieved latency of selected representative networks as measured across the evaluated platforms. To investigate the trade-off between accuracy and achieved processing speed, we analyze the inference latency (\textit{i.e.}, batch size of 1) against the accuracy of these networks. The results (Figure \ref{fig:inf_per_acc}) demonstrate that there are significant differences among the evaluated hardware platforms. 

In the workstation setting, we observe that the server-grade CPU is sub-optimal when it comes to supporting very deep networks (\textit{e.g.}, the ResNet family) or those with large number of FLOPs (\textit{e.g.}, VGG). As a result, none of the state-of-the-art networks could sustain more than 15 FPS. Instead, the RTX 2080 Ti achieves a throughput improvement over the Xeon 4116 CPU in the range $7\times$-$50\times$  with an average of 15$\times$ across the networks. In particular, networks such as ResNet yield a speedup of $\approx14\times$, whereas simpler networks such as AlexNet can handle an impressive 680 FPS. Similarly, with respect to power efficiency, RTX 2080 Ti yields an average improvement of 5.3$\times$ in inferences/s/W over the Xeon CPU.

In the <30 watts range, the Tegra Xavier GPU manages to outperform the Xeon CPU with raw speedups between $2\times$-$15\times$ (5.1$\times$ average) and yields an average power efficiency gain of 14.4$\times$ across the networks. Compared to the RTX 2080 Ti GPU, although the Tegra Xavier GPU reaches between $26$-$37\%$ ($34\%$ average) of its raw performance, it achieves a 2.8$\times$ average improvement in power efficiency, demonstrating its suitability for applications when inferences/s/W are the primary metric of interest.

Due to their resource constraints, the Kryo mobile CPU exhibits substantially lower raw performance when compared to the server-grade CPU and GPU platforms; NCS 2 can handle up to 23 FPS on average whereas the Kryo processor achieves less than 6 FPS (35$\times$ slower than the RTX 2080 Ti GPU). Despite the expected degradation of the mobile CPU in terms of raw throughput, NCS 2 achieves a power efficiency in the range $3\times$-$88\times$ (41.6$\times$) over the power-hungry RTX 2080 Ti GPU. In this respect, the NCS 2 platform extracts the maximum performance out of its 1-watt TDP and constitutes a powerful candidate device for performing DNN inference in very-low-power settings.

Interestingly, by observing the accuracy-latency Pareto fronts in Figure \ref{fig:inf_per_acc}, mobile devices demonstrate substantially different patterns compared to the server-grade experiments. First, deeper networks with large number of operations suffer from a big penalty in raw performance. For instance, ResNet on the mobile CPU results in a minimal throughput of 0.2 inferences-per-second which is 250$\times$ slower than the RTX 2080 Ti GPU. At the same time, networks that are optimized for mobile devices do significantly better: MobileNetV2 is 6.4$\times$ faster on NCS 2 and 16.3$\times$ faster on the Kryo processor compared to ResNet. Despite the accuracy penalty, the design decision of using depthwise separable convolutions seems to offer a considerable performance benefit in the mobile space. At the same level of accuracy, VGG16 is uniformly slower across devices by a considerable margin (ranging from 2$\times$ to 6$\times$).
Finally, ShuffleNetV2 is a notable example that achieved significantly improved performance on Kryo CPU when compared to server-grade runs, demonstrating the benefits of pointwise group convolution and channel shuffling for resource-constrained devices. 

Overall, from a high-level view, the Pareto fronts of low-power devices (\textit{i.e.}, sub-figures on second row of Figure \ref{fig:inf_per_acc}) comprise different networks compared to their more power-consuming counterparts. The Pareto front of the Xeon 4116 CPU comprises AlexNet, MobileNetV2 and ResNet152, while the RTX 2080 Ti and Tegra Xavier GPUs contain AlexNet, VGG16 and ResNet152. On the other hand, NCS 2 also includes DenseNet201 on its Pareto front, with VGG16 being excluded due to its very latency-expensive NCS 2 mapping, while ShuffleNetV2 and NASNet-mobile also appear as Pareto optimal networks for the CPU of SDM845. As a result, the direct use of hardware-agnostic metrics, such as number of FLOPs, can often be misleading and not accurately indicate how efficiently a DNN is mapped on a particular platform. 

\textbf{Per-layer analysis.}
To further investigate into why some networks are mapped more efficiently on certain platforms, we also look into the breakdown of inference time within each operation (Figure \ref{fig:per_layer}). As already demonstrated in the literature, the majority of time is spent on convolution operations ranging from 65\% of the time in desktop GPU to 89\% of the time on mobile processors. This difference further demonstrates the need for optimizing these operations on mobile platforms. For NVIDIA GPU accelerators, the second most time-consuming operation for this workload was Batch Normalization (19\% and 15\% of time on the RTX 2080 Ti and Xavier GPUs respectively) whereas for the Xeon CPU Max Pool becomes substantial by occupying 10\% of the time. Finally, on the Kryo mobile processor 10\% of the time is spend on fully-connected layers. AlexNet fully reveals that fully-connected layers are not a good fit for the characteristics of mobile platforms, taking more than 70\% of the computation time when compared to below 30\% on server-grade GPUs. The primary factor behind the slow execution of fully-connected layers is the limited off-chip memory bandwidth of existing mobile platforms which determines the processing speed of the inherently memory-bound operations of fully-connected layers.

\vspace{-0.25cm}
\section{Discussion}
The analysis of Section \ref{sec:evaluation} uncovers a number of notable insights.

\textbf{Performance variability:} We note that the performance of each network varies substantially across platforms depending on the network architecture and the type of operations used. Therefore, an interesting research direction would be to design tools that can automatically select when to use and fuse together these building blocks, depending on the hardware architecture of the target device as well as the latency and throughput requirements \cite{cai2018proxylessnas,Crowley2017,ashok2018nn}. 

\textbf{Mobile-specific optimizations:} Our results further demonstrate the importance of mobile-specific operations such as pointwise and depthwise separable convolutions. Furthermore, our benchmarks show that mobile devices are inefficient in handling larger models due to their reduced memory capabilities. Therefore, while compression and quantization techniques might not result in a big performance gain on desktop environments, they do make a big difference on mobile and embedded devices, both compute- and memory-wise.
Towards this direction, binary networks \cite{DBLP:journals/corr/CourbariauxB16,Rastegari2016} seem to offer a promising alternative for maximal compression, but currently require specialized hardware support to exploit the speedup potential.
Finally, it is possible to dynamically offload computation from device to the edge or cloud, in order to facilitate computation and minimize latency \cite{Cartas:2019:RCI:3301418.3313946}.

\textbf{Importance of hardware support:} Most of the examined mobile devices either do not come with automated libraries for targeting their accelerator backends (GPU, DSP, NPU), or provide limited support for DNN operators.
Our results demonstrate that most of the time is consumed on operations such as convolutions and fully-connected layers. Instead of being limited to the CPUs, software optimizations and support for exploiting the available hardware accelerators of target mobile and embedded chipsets should be prioritized in order to accelerate these common operations in a transparent and homogeneous way \cite{facebook_edge_2019}.

\textbf{Batch size:} While most devices are optimized for larger batch sizes, most real-time mobile applications require smaller batch sizes to minimize latency. We observe that across all platforms the hardware is not fully utilized for smaller batch sizes.
One possible direction is to study how multiple networks with small batch sizes can be optimally collocated to overprovision these resources and thus maximize their utilization \cite{Jeon,Narayanan}.

\section{Conclusions}
\label{sec:concolusions}

In this work, we attempted to shed some light on the performance of DNN inference by analyzing more than 25 networks on a wide variety of commodity devices. 
Our results provide useful insights about the performance and suitability of these models. 
Furthermore, we identified model design choices that work better on each platform and we uncovered a number of performance bottlenecks.
We believe that these results can help the research community in three ways: i) to identify the best possible building blocks when designing models for these platforms, ii) shed light on the capabilities of these devices and iii) provide insights about possible future hardware and software optimizations.

\vspace{+0.2cm}
\bibliographystyle{plain}
\bibliography{references}
\end{document}